# Detecting Small Signs from Large Images

Zibo Meng*, Xiaochuan Fan†, Xin Chen†, Min Chen‡ and Yan Tong*
* Computer Science and Engineering
University of South Carolina, Columbia, SC USA
Email: mengz@email.sc.edu, tongy@cec.sc.edu
†HERE North America, Chicago, IL USA
Email: efan3000@gmail.com, xin.5.chen@here.com
‡Computing and Software Systems, School of STEM
University of Washington Bothell, Bothell, WA USA
Email: minchen2@uw.edu

*Abstract*—In the past decade, Convolutional Neural Networks (CNNs) have been demonstrated successful for object detections. However, the size of network input is limited by the amount of memory available on GPUs. Moreover, performance degrades when detecting small objects. To alleviate the memory usage and improve the performance of detecting small traffic signs, we proposed an approach for detecting small traffic signs from large images under real world conditions. In particular, large images are broken into small patches as input to a Small-Object-Sensitive-CNN (SOS-CNN) modified from a Single Shot Multibox Detector (SSD) framework with a VGG-16 network as the base network to produce patch-level object detection results. Scale invariance is achieved by applying the SOS-CNN on an image pyramid. Then, image-level object detection is obtained by projecting all the patch-level detection results to the image at the original scale. Experimental results on a real-world conditioned traffic sign dataset have demonstrated the effectiveness of the proposed method in terms of detection accuracy and recall, especially for those with small sizes.

## I. INTRODUCTION

Object detection is an important task in computer vision for computers to understand the world and make reactions, and has great potential to emerging applications such as automatic driving. In the past few years, deep Convolutional Neural Networks (CNNs) have shown promising results on object detection [1], [2], [3], [4], [5]. The existing systems can be divided into proposal-based methods, such as R-CNN [1], Fast-RCNN [2], Faster-RCNN [3], and proposal-free methods, such as You Only Look Once (YOLO) [4] and Single Shot Multibox Detector (SSD) [5].

Although CNNs have been demonstrated to be effective on object detection, existing methods often cannot detect small objects as well as they do for the large objects [5]. Moreover, the size of input for those networks is limited by the amount of memory available on GPUs due to the huge memory requirements for running the network. For example, an SSD object detection system [5] based on VGG-16 [6] requires over 10 gigabytes taking a single image with a size of 2048 × 2048 as input. One way to overcome the aforementioned problem is to simplify the network, e.g. using a shallow one, with a tradeoff of performance degradation. A second possible solution is to down-sample the original image to fit the memory. However, the small objects will be even more difficult to detect.

In this work, we propose a novel approach to address the aforementioned challenges for accurate small object detection from large images (e.g. with a resolution of over 2000 × 2000). As shown in Fig. 1, to alleviate the large memory usage, the large input image is broken into patches with fixed size, which are fed into a Small-Object-Sensitive convolutional neural network (SOS-CNN) as input. Moreover, since large objects may not be entirely covered in a single image patch, the original image is down-sampled to form an image pyramid, which allows the proposed framework to achieve scale-invariance and to process input images with a variety of resolutions.

As illustrated in Fig. 2, the proposed SOS-CNN, employs a truncated SSD framework using a VGG-16 network as the base network, where only the first 4 convolutional stages are kept. SSD [5] used a number of default boxes and a set of convolutional layers to make predictions on multiple feature maps with different scales. The network becomes deeper as the input gets larger since extra convolutional layers are required to produce predictions of larger scales. Different from SSD, the proposed method makes predictions on only one feature map of small scale and achieves scale invariant by constructing an image pyramid.

To sum up, our contributions are as follows:
- An object detection framework, which is capable of detecting small objects from large images, is introduced.
- An SOS-CNN, which is sensitive to small objects, is designed to improve the performance on small object detection in large images.

Most of the current object detection datasets, e.g. PASCAL VOC [7] and ImageNet ILSVRC [8], contain images where the objects have proper sizes and have been well aligned. However, this is not the case in the real world, where most of the objects occupy a small portion of the whole image, and sparsely distributed. Thus, in this paper, a sign detection database [9] consisting of images collected under real world conditions is employed to evaluate the proposed approach. Experiments on the real-world traffic sign detection database have demonstrated the effectiveness of the proposed framework especially for small signs.

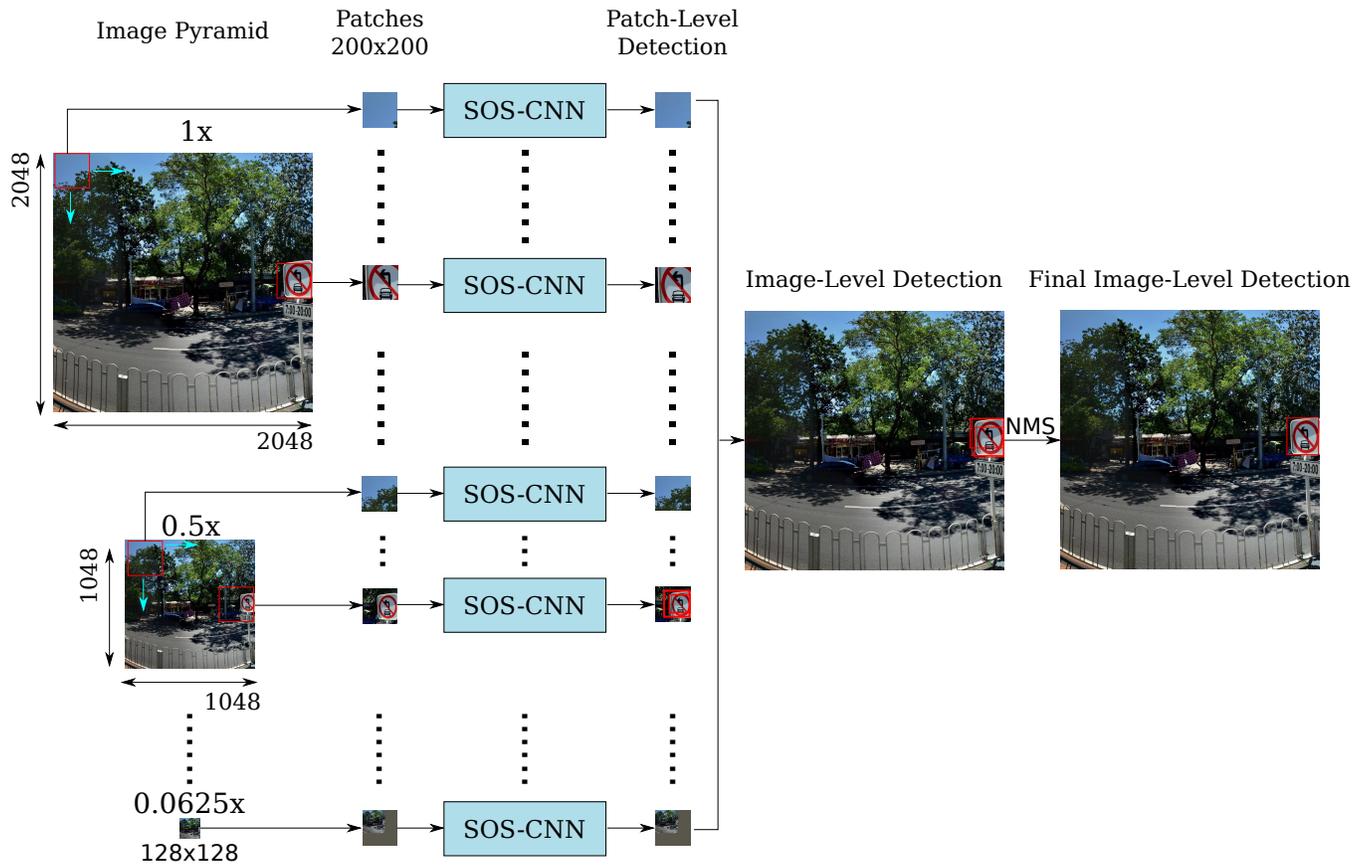

Figure 1. An illustration of the proposed framework. The original image is broken into small patches with fixed sizes as input to a SOS-CNN to produce patch-level object detection results. Moreover, an image pyramid is built, on which the proposed SOS-CNN will be applied, to achieve scale invariance. Image-level object detection results are produced by projecting all the patch-level results back to the original image. Non-maximum suppression (NMS) is employed to generate the final image-level prediction. The whole process can be done in an end-to-end fashion.

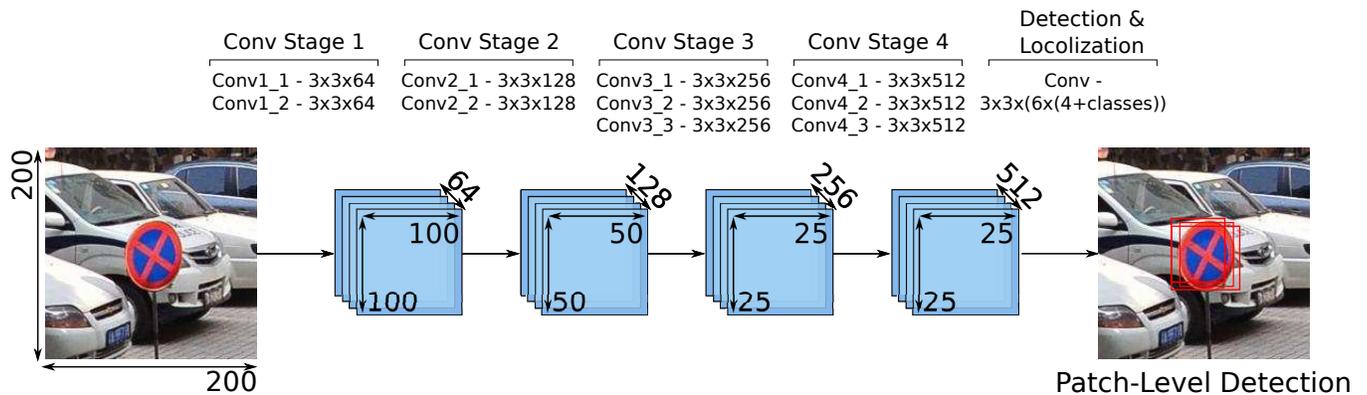

Figure 2. An illustration of the proposed SOS-CNN. Taking $200 \times 200$ images as input, a truncated SSD framework with a VGG-16 network as the base network is employed to produce patch-level detection, where only the first 4 convolutional stages are kept. A set of convolutional layers with a kernel size of $3 \times 3$ are appended to the end of the network for object detection and localization.

## II. RELATED WORK

As elaborated in survey papers [10], object detection has been extensively studied in the past decades. Deformable Part Model (DPM) [11] and Selective Search [12] have shown promising results on object detection. Recently, deep learning has been demonstrated in object detection [1], [2], [3], [4], [5].

Existing deep learning based practice can be divided into two categories: proposal-based and proposal-free methods. Among the proposal-based approaches, R-CNN [1] is the first deep learning approach improving the object detection performance by a large margin. R-CNN generates proposals, i.e., candidate regions for object detection, using an object proposal algorithm [13], produces features for each proposal region using a CNN, and makes final predictions using an SVM. Since then, several approaches based on R-CNN have been developed. SPPnet [14] speeds up the detection speed of R-CNN by introducing a spatial pyramid pooling (SPP) layer, which shares features between proposals, and is more robust to image size and scale. Based on SPP layer, Fast-RCNN [2] makes the network can be trained end-to-end. Faster-RCNN [3] proposes a region proposal CNN and integrates it with Fast-RCNN by sharing convolutional layers, which further improves the object detection performance in terms of detection speed and accuracy. Most recently, many systems are developed following the diagram established by Faster-RCNN [15], [16] and have achieved promising results on object detection.

Another stream contains proposal-free methods, where the object detection results are produced without generating any object proposals. OverFeat method [17] presents an approach to predict the class label and the bounding box coordinates by applying a sliding window on the top-most feature map. YOLO [4] uses a fully connected layer to produce categorical predictions as well as bounding box coordinates simultaneously on the top-most feature map. SSD [5] uses a set of convolutional layers as well as a group of pre-defined prior boxes to make predictions for each location on multiple feature maps of different scales. The proposal-free approaches can generally produce comparable accuracy compared with proposal-based methods while having faster detection speed by discarding an additional object proposal generation step.

However, the computation capacity of all the current deep learning based methods is limited by the memory available on GPUs. It becomes infeasible to use a deep CNN based approach to process a large image, e.g. with a size of $2048 \times 2048$. In addition, we designed an proposal-free SOS-CNN to improve the performance on small object detection.

Different from the previous approaches that directly employ the original images as input, we break the large images into small patches to alleviate the memory requirement when the size of the input is large. Moreover, an image pyramid is created to achieve scale invariance. Our framework can be trained end-to-end using standard stochastic gradient descent (SGD).

## III. METHODOLOGY

In this section, we first give the overview of the proposed model for detecting small objects from large image in Sec III-A. The details for the training process and the testing process will be given in Sec III-B and Sec III-C, respectively.

### A. Overview

The proposed framework employs an image pyramid, each of which is broken into patches with the same size as input to the SOS-CNN to produce patch-level detection. Non-maximum suppression (NMS) is employed to generate the final predictions on the original image.

*1) Multi-patch detection:* Since the memory available on GPUs is limited, the VGG-16 network cannot process large images, i.e. with a size of $2048 \times 2048$. To alleviate the memory demand, small patches with fixed size, i.e. $W \times H$, will be cropped from the large images as the input to the SOS-CNN. The patches are obtained in a sliding window fashion with a stride of $s$ in both horizontal and vertical direction on the large image.

*2) Scale Invariant Approach:* Since the SOS-CNN is designed to be sensitive to small objects, objects with larger sizes will not be detected in the original image. Thus, a scale invariant detection approach is developed. Particularly, given an input image, an image pyramid is constructed, where the larger objects that cannot be captured in the image with original resolution become detectable on images with smaller scales.

*3) SOS-CNN:* The proposed framework includes an SOS-CNN which is designed for small object detection. As illustrated in Fig 2, the network is derived from an SSD model [5] with a VGG-16 network. A set of convolutional layers with $3 \times 3$ kernels are employed to produce the confidence scores for each category as well as the offsets relative to a group of pre-defined default boxes for each location on the top-most feature map, i.e. the output feature map of $Conv4_3$.

*Single scale feature map for detection:* In this work, we produce the object detection on the feature map generated by the top-most feature map, i.e. *conv4_3* in Fig. 2. The receptive field of this layer is $97 \times 97$, which is adequate for small object detection, yet can offer some context information, which has been proven crucial for small object detection [18], [19].

*Default boxes and aspect ratios:* Similar to the approach in Faster-RCNN [3] and SSD [5], a set of pre-defined default boxes with different sizes and aspect ratios are introduced at each location of the top-most feature map to assist producing the predictions for bounding boxes. Instead of directly predicting the location of the bounding boxes for each object in an image, for each position of the feature map, the SOS-CNN predicts the offsets relative to each of the default boxes and the corresponding confidence scores over the target classes simultaneously. Specifically, given $n$ default boxes associated with each location on the top-most feature map with a size of $w \times h$, there are $n \times w \times h$ default boxes in total. For each of the default boxes, $c$ classes and 4 offsets relative to the default box location should be computed. As a result, $(c+4) \times n \times w \times h$ predictions are generated for the feature map.

To sum up, the proposed framework uses a single feature map of small scale, while achieving scale-invariance by manipulating scale of inputs, so that the network can focus on learning the discriminative features for small objects while being invariant to scale differences.

### B. Training

*1) Data Preparation:* $200 \times 200$ patches centered at target objects are cropped from the original images as input of the network. There are two cases we should consider. First, the target objects may be larger than the patch at the current pyramid level. Second, multiple objects might be included in one patch. An object is labeled as positive only if over 1/2 area of the object is covered in the patch. In addition, to include more background information, a set of patches containing only background are randomly cropped from the original training images for learning the model. The ratio between the number of background patches and that of the positive patches is roughly 2:1.

*2) Choosing sizes and aspect ratios of the default boxes:* To ensure the network focusing on detecting small objects, default boxes with small sizes are chosen. In particular, given the input size of the network as $200 \times 200$, the size of the square default boxes are $\mathcal{S}_1 = 0.1 \times 200$, and $\mathcal{S}_2 = \sqrt{(0.1 \times 200) \times (0.2 \times 200)}$, which means the model will focus on the objects that occupy around 10% of area of the input image. To make the model fit better to objects with a shape other than square, different aspect ratios are chosen for the prior boxes, which are $\mathcal{R} \in \{2, 3, \frac{1}{2}, \frac{1}{3}\}$. Given the aspect ratio, $\mathcal{R}$, the width, i.e. $w_\mathcal{R}$, and height, $h_\mathcal{R}$ of the corresponding default box can be calculated as:

$$w_\mathcal{R} = \mathcal{S}_1 \sqrt{\mathcal{R}}$$
$$h_\mathcal{R} = \frac{\mathcal{S}_1}{\sqrt{\mathcal{R}}}$$

As a result, there are 6 default boxes associated with each cell on the top-most feature map of the SOS-CNN with a size of $25 \times 25$. Given scores over $c$ classes and 4 offsets relative to each box for each location on the feature map, $(c+4) \times 6 \times 25 \times 25$ predictions are generated for each input image.

*3) Matching Default Boxes:* During training stage, the correspondence between the default boxes and the ground truth bounding boxes is firstly established. In particular, the Jarccard overlap between each default box and the ground truth boxes are calculated. The default boxes are considered as "matched" when the Jaccard overlap is over 0.5. As illustrated in Fig. 3, the object at original size is too large to be matched by the default boxes, as illustrated in Fig. 3(a), where the solid blue rectangle gives the ground truth box, since the default boxes are designed to be sensitive to only objects with small sizes. After being down-sampled 3 times, the objects becomes matchable in the down-sampled image, as shown in Fig. 3(b). Analogous to regressing multiple boxes at each location in YOLO [4], different default boxes can be matched to one ground truth box, as depicted in Fig. 3(c), where the dashed blue rectangles represent the default boxes matched with the ground truth, while the dashed gray rectangles give the unmatched boxes. For each of the matched boxes, offsets relative to the box shape and the corresponding confidence scores are produced, as depicted in Fig. 3(c), which are used to calculate the loss and update the parameters of SOS-CNN.

*4) Objective Function:* The proposed SOS-CNN employs an objective function to minimize localization loss and classification loss [5], which is defined as follows:

$$\mathcal{L}(x, y, \hat{b}, b) = \frac{1}{N} \left( \mathcal{L}_{conf}(x, y) + \lambda \mathcal{L}_{loc}(x, \hat{b}, b) \right) \quad (1)$$

where $x$ is a matched default box; $N$ is the number of matched default boxes, and $\mathcal{L}_{loc}(\cdot)$ is the Smooth L1 loss [2] based on the predicted box, i.e. $\hat{b}$ and the ground truth bounding box, i.e. $b$; $\mathcal{L}_{conf}(\cdot)$ is the softmax loss over target classes; and $\lambda$ is the weight to balance between the two losses, which is set to 1 in our experiment empirically.

*5) Data Augmentation:* To make the model more robust to input object shape and location differences, similar data augmentation approach is employed as in SSD [5], where training samples will be produced by cropping patches from the input images. The overlapped part of the ground truth box will be kept if over 70 percent of its area falls in the sampled patch. The sampled patch is resized to a fixed size, i.e. $200 \times 200$, as input for training the SOS-CNN.

*6) Hard Negative Sampling:* Hard negative samples are selected for training according to the confidence scores after each iteration during the training process. In particular, at the end of each training iteration, the miss-classified negative samples will be sorted based on the confidence scores and the ones with the highest confidence scores will be considered as hard negative samples, which are used to update the weights of the network. Following the implementation in SSD [5], the number of hard negatives

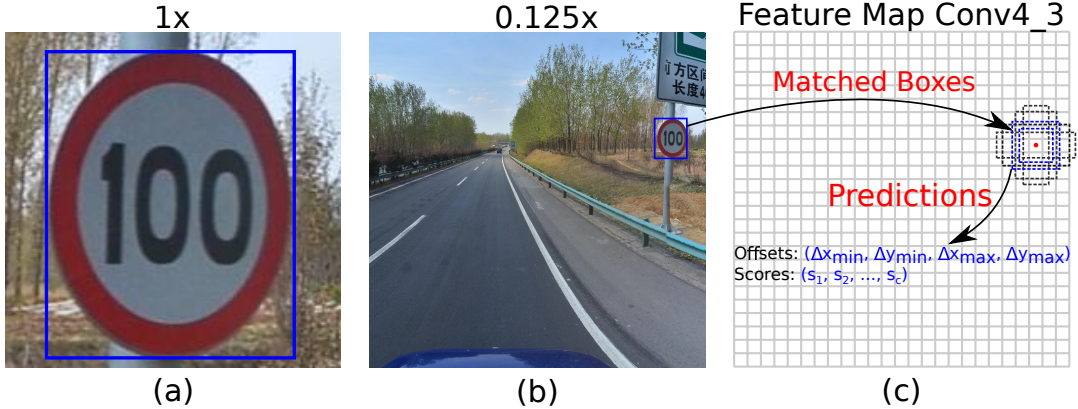

Figure 3. An illustration of the matching process during training stage. A set of default boxes is assigned to each location on the feature map, as depicted in (c), and the default box is considered as "matched" if the Jaccard overlap between it and the ground truth bounding box is over 0.5. The object with original size can not be matched with any default boxes, as illustrated in (a), where the solid blue rectangle gives the ground truth box, since the size is larger than that of the designed default boxes. After being down-sampled 3 times, the objects becomes "matchable", as shown in (b), and can be matched multiple times with the default boxes on the $25 \times 25$ feature map, as depicted in (c), where the dashed blue rectangles are the matched prior boxes, and the dashed gray rectangles are the default boxes that cannot be matched. For each of the matched box, offsets relative to the box shape and corresponding confidence scores are produced.

used for training the model is at most 3 times larger than that of positives.

### C. Testing

*1) Multi-patch Testing:* Since the limited amount of memory available on current GPUs, it is infeasible for deep networks to accept large images as input, i.e. with a size of $2048 \times 2048$. Thus, $200 \times 200$ patches will be cropped from the input image, which will be fed into the trained SOS-CNN for testing.

*2) Multi-scale Testing:* As the SOS-CNN is designed to be sensitive to small objects, some large signs in the original image will be missed at the original resolution. An image pyramid is created to cope with the problem. Specifically, as illustrated by the left most column in Fig. 1, given an input image, a smaller image is obtained by sub-sampling the input image by a factor of $r$ along each coordinate direction. The sample procedure is repeated several times until a stop criterion is met. $200 \times 200$ patches are cropped from each of the images in the pyramid, which are employed as input to the SOS-CNN to produce patch-level detection. Image-level detection can be obtained by utilizing NMS. The image pyramid constructing and patch-cropping process can be done on-the-fly during the testing process.

*3) Multi-batch Testing:* It is impossible to put all the patches from a single image into one testing batch because of the memory limitation on current GPUs. Thus, we design a process to divide the patches from the same image into several batches. All the patch-level predictions will be projected back onto the image at the original scale after all the patches from the same image are processed. Then NMS is employed to generate the final image-level predictions as illustrated in Fig 1.

### IV. EXPERIMENTAL RESULTS
### A. Implementation Details

The SOS-CNN is trained by using an initial learning rate of 0.001, which is decreased to 0.0001 after 40,000 iterations, and continues training for another 30,000 iterations. A momentum of 0.9 and a weight decay of 0.0005 are employed.

During testing, an image pyramid will be constructed with a down-sampling ratio $r = 0.5$, until the area of the down-sampled image falls below 0.4 of $200 \times 200$. $200 \times 200$ patches are cropped from each of the images in the pyramid with a stride of $s = 180$ in both horizontal and vertical directions. The last part in the horizontal direction will be padded by zeros if it does not fit the patch completely. The last part in the vertical direction gets discarded if it does not make a whole patch.

When evaluating the results, we use a threshold of $0.5$ for the confidence score and an intersection over union (IoU) of $0.5$ between the predicted bounding box and ground truth. The proposed method is implemented in CAFFE library [20] and trained using SGD.

### B. Tsinghua Traffic Sign Detection Dataset

The Tsinghua traffic sign detection database [9] is composed of 10,000 images containing 100 classes of traffic signs with a resolution of $2048 \times 2048$. The images are collected under real world conditions with large illumination variations and weather differences. Each traffic sign instance generally occupies only a small proportion of an image, e.g. 1%. The database comes with training and testing sets partitioned, while the categorical labels as well as the bounding box associated with each sign are given. The ratio of the numbers of images in training and testing sets is roughly 2, which is designed to offer enough variations for training a deep model.

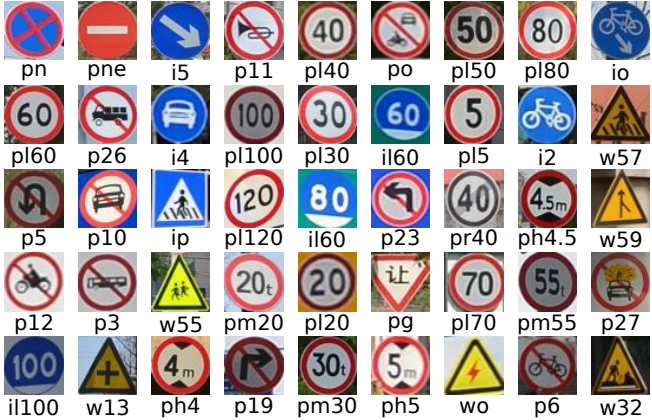

Figure 4. Examples of the 45 classes of traffic signs and their notations used in the experiment from Tsinghua traffic sign detection database.

## C. Results on Tsinghua Traffic Sign Detection Dataset

*1) Data Preparation:* Following the configurations in [9], only traffic signs of 45 classes, whose numbers of instances in the dataset are larger than 100, are selected for evaluating the proposed framework. Examples of signs selected for experiment in this work as well as their notations are shown in Fig. 4. The data is prepared for training as described in Sec. III-B1.

*2) Experimental Results:* To better demonstrate the effectiveness of the proposed method on small sign detection while maintaining the power for detecting objects with larger sizes, the signs are divided into three different groups according to their areas, i.e. small ($Area \in [0, 32^2]$), medium ($Area \in (32^2, 96^2]$), and large ($Area \in (96^2, 400^2]$). Note that, even signs falling in the large group has relatively small size compared to the size of the original image, i.e. a sign with a size of $400 \times 400$ occupying about 3.8% area of the original image ($2048 \times 2048$).

The recall-accuracy curves for two state-of-the-art methods, i.e., Fast-RCNN and Zhu et al. [9], and the proposed approach are plotted in Fig. 5. The curves for Fast-RCNN and Zhu et al. are adopted from [9]. Note that the Fast-RCNN employed VGG_CNN_M_1024 [21] as the base network, which employs a large stride on the first convolutional layer to be able to process the large images. For the proposed framework, the accuracy-recall curve is produced using all the predictions with a confidence score above 0.01. The proposed method consistently outperforms the two state-of-the-art methods on signs of different sizes. More importantly, the proposed system outperforms Fast-RCNN on the small signs by a large margin, indicating the effectiveness of the proposed method on small sign detection. Overall, Fast-RCNN has a recall of 0.56 and an accuracy of 0.50, Zhu et al. achieved a recall of 0.91 and an accuracy 0.88, while our approach has a recall of 0.93 and an accuracy of 0.90.

*3) Discussion:* To demonstrate that the proposed framework is sensitive to small objects and scale invariant, we conducted another three experiments on different testing data:

- Using the patches only from the images with original resolution, i.e. $2048 \times 2048$, as input to the SOS-CNN without any down-sampling process for testing, denoted as "*High*" for high resolution;
- Using the patches from the image that has been down-sampled once, i.e. $1024 \times 1024$, as input without any further resizing, which is denoted as "*Medium*";
- Using the patches from the image that has been down-sampled twice, i.e. $512 \times 512$, and those from the images that have been down-sampled until the stop criterion is met, which is denoted as "*Low*".

The results on the Tsinghua traffic sign detection dataset of the three experiments are depicted in Fig. 6. On the image with high resolution, i.e. original images with a resolution of $2048 \times 2048$, since the network is designed to be sensitive to the small objects, the detection performance for "*High*" on signs with small sizes is the best, i.e. blue curve in Fig. 6(a), compared with that for signs with medium and large sizes, i.e. blue curves in Fig. 6(b), and (c). On the image with low resolution, where the originally large signs become detectable by the SOS-CNN, while the originally small signs become invisible to the network, the detection performance for "*Low*" on large signs, i.e. green curve in Fig. 6(c), becomes superior to that on the images with high or medium resolutions, i.e. green curves in Fig. 6(a), and (b). For the signs whose size falls in $(32, 96]$, some of them can be well captured in the original image and some of them will become detectable after down-sampling once, as illustrated in 6(b), "*High*" and "*Medium*" both perform reasonably well, i.e. blue and red curves in Fig. 6(b), respectively, since they can predict part of the signs with medium sizes. By combining the results from images with different resolutions, the proposed method becomes scale invariant and achieves better performance on signs with different sizes compared with state-of-the-arts as illustrated in Sec. IV-C2

## V. CONCLUSION AND FUTURE WORK

In this work, a framework for detection small objects from large image is presented. In particular, due to the limited memory available on current GPUs, it is hard for CNNs to process large images, e.g. $2048 \times 2048$, and even more difficult to detect small objects from large images. To address the above challenges, the large input image is broken into small patches with fixed size, which are employed as input to an SOS-CNN. Moreover, since objects with large sizes may not be detected in the original resolution, an image pyramid is constructed by down-sampling the original image to make the large objects detectable by the SOS-CNN. The SOS-CNN is derived from an SSD model with a VGG-16 network as the base network, where only the first 4 convolutional stages of VGG-16 network are kept. A group of default boxes are associated with each location on the feature map to assist the SOS-CNN to produce object detection. A set of convolutional layers with a kernel size of $3 \times 3$ is employed to produce the confidence scores and coordinates of the corresponding bounding box for each of the default boxes. Experimental results on a traffic sign detection dataset, which includes images collected under real world conditions, containing signs occupying only a small proportion of an image, have demonstrated the effectiveness of the proposed method in terms of alleviating the memory usage while maintaining a good sign detection performance, especially for signs with small sizes.

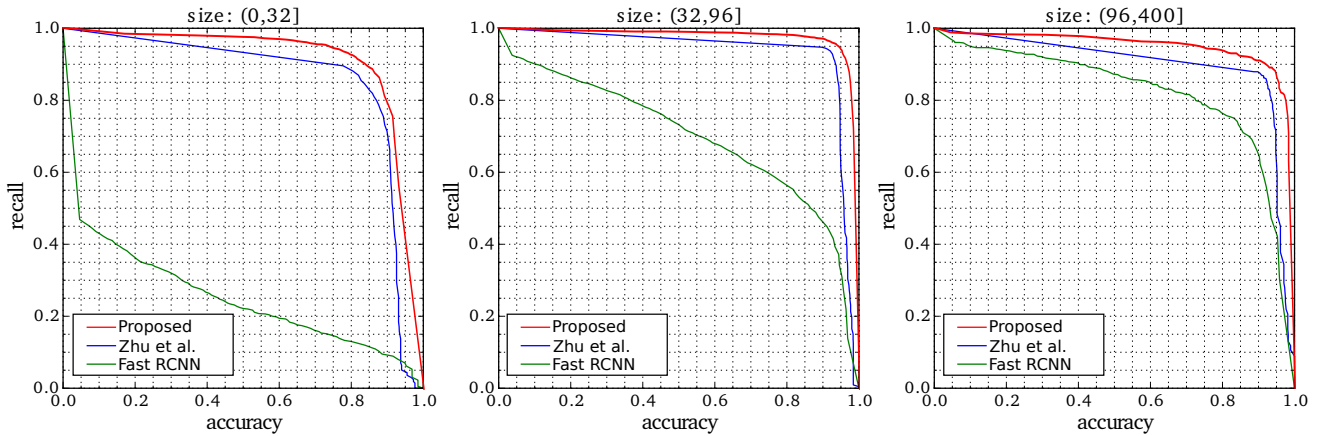

Figure 5. Performance comparison on the Tsinghua traffic sign detection database for small, medium, and large signs. The accuracy-recall curves for Fast-RCNN and Zhu et al. are adopted from [9], and the one for the proposed method is produced using all the detection results with a confidence score above 0.01. The proposed method consistently outperforms Fast-RCNN and the method by Zhu et al. on signs for all different sizes. The performance is more impressive on small signs compared to Fast-RCNN.

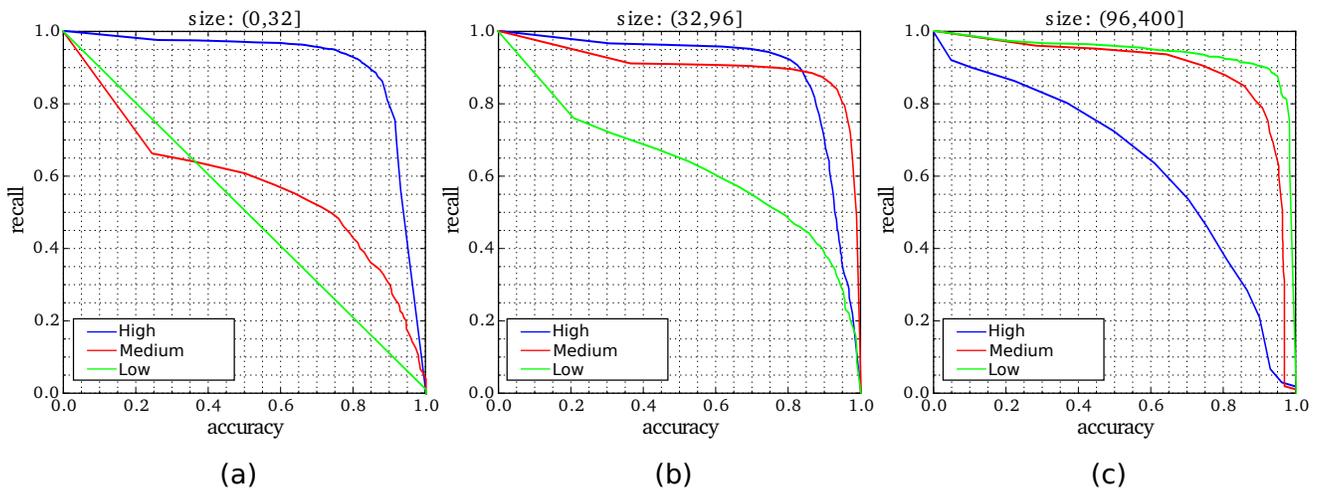

Figure 6. An illustration of the effectiveness of the proposed SOS-CNN in terms of detecting small signs. On the images with high resolution, i.e. $2048 \times 2048$, the detection performance for signs with small sizes, represented by the blue curve in (a), is the best compared with that for signs with medium and large sizes in (b) and (c). On the images with low resolutions, i.e. less than or equal to $512 \times 512$, the originally large signs become detectable by the SOS-CNN, and thus the detection performance for the large signs, denoted by the green curve in (c), becomes superior to that on the images with high or medium resolutions in (a) and (b).

Since the proposed system employed a sliding window strategy, it is time consuming. In the future, we plan to make the system more efficient.